\documentclass{article}

\usepackage{arxiv}

\usepackage{amsmath}
\usepackage[utf8]{inputenc} 
\usepackage[T1]{fontenc}    
\usepackage{booktabs}       
\usepackage{amsfonts}       
\usepackage{nicefrac}       
\usepackage{microtype}      
\usepackage{lipsum}         
\usepackage{graphicx}
\usepackage{natbib}

\usepackage{graphicx}

\usepackage{caption}
\usepackage{multirow}
\usepackage{algorithm}
\usepackage{algorithmic}
\usepackage{array}
\usepackage{threeparttable}
\usepackage[caption=false,subrefformat=parens]{subfig}
\usepackage{marvosym}
\usepackage[normalem]{ulem}

\DeclareSubrefFormat{parens}{#1(#2)}

\newcommand\blfootnote[1]{%
  \begingroup
  \renewcommand\thefootnote{}\footnote{#1}%
  \addtocounter{footnote}{-1}%
  \endgroup
}

\title{Graph Augmentation Learning}

\author{Shuo Yu\\
		School of Software\\
		Dalian University of Technology\\		
		Dalian, Liaoning, China, 116620\\
		shuo.yu@ieee.org
		\And
		Huafei Huang\\
		School of Software\\
		Dalian University of Technology\\		
		Dalian, Liaoning, China, 116620\\
		hhuafei@outlook.com
		\And
		Minh N. Dao\\
		School of Engineering, IT and Physical Sciences\\
		Federation University Australia\\
		Ballarat, VIC, Australia, 3353\\
		m.dao@federation.edu.au
		\And
		Feng Xia\thanks{Corresponding author.}\\
		School of Engineering, IT and Physical Sciences\\
		Federation University Australia\\
		Ballarat, VIC, Australia, 3353\\
		f.xia@ieee.org
		}

\begin{document}
\maketitle

\begin{abstract}
	Graph Augmentation Learning (GAL) provides outstanding solutions for graph learning in handling incomplete data, noise data, etc. 
  Numerous GAL methods have been proposed for graph-based applications such as social network analysis and traffic flow forecasting. 
  However, the underlying reasons for the effectiveness of these GAL methods are still unclear. As a consequence, how to choose optimal graph augmentation strategy for a certain application scenario is still in black box. 
  There is a lack of systematic, comprehensive, and experimentally validated guideline of GAL for scholars. 
  Therefore, in this survey, we in-depth review GAL techniques from macro (graph), meso (subgraph), and micro (node/edge) levels. 
  We further detailedly illustrate how GAL enhance the data quality and the model performance. The aggregation mechanism of augmentation strategies and graph learning models are also discussed by different application scenarios, i.e., data-specific, model-specific, and hybrid scenarios. 
  To better show the outperformance of GAL, we experimentally validate the effectiveness and adaptability of different GAL strategies in different downstream tasks. 
  Finally, we share our insights on several open issues of GAL, including heterogeneity, spatio-temporal dynamics, scalability, and generalization.\blfootnote{This work was accepted in The First International Workshop on Graph Learning in IW3C2.}
\end{abstract}


\section{Introduction}
\label{sec:intro}

Graph structured data are everywhere. In real world, analyzing graph data has numerous applications and significantly improves human daily life~\cite{Xia2021TAI,Yu2021TCSS}. Car-hailing applications analyze traffic network data with graph-based methods to better predict traffic flow~\cite{DBLP:journals/isci/TaoWGH17}. Social media platforms provide personalized recommendation content to different users~\cite{DBLP:journals/tkde/JiangC16,DBLP:conf/www/ZhangGPHFYL019}. Graph-based approaches are designed to optimize the efficiency of resource scheduling in the Internet of Things (IoT)~\cite{DBLP:conf/infocom/JiaXYCLW18}. Academic networks are analyzed to better enhance the collaboration among scientists~\cite{Kong2019JNCAacademic,DBLP:conf/sigir/0006RZCL020}. 
However, real-world graph data are always incomplete due to many reasons such as privacy policies and loss in data collection. Many graph learning methods lack the ability in handling this, thus these methods are generally proposed and verified based on the assumption that data are of high quality. Such neglect leads to suboptimal or even wrong results. 
Some graph learning methods are designed for special application scenarios that have to consider the data quality in front, such as advisor-advisee recognition~\cite{liu2019TKDE} and fake review detection~\cite{DBLP:conf/incdm/CaoLYC21}. 
However, these customized graph learning methods can rarely be applied in other tasks. Moreover, redesigning algorithms for another task or application is generally time-consuming, high cost, and there is no guarantee that these limitations can be overcome. 

Graph augmentation approaches have been demonstrated to be effective in addressing restrictions in graph learning, e.g., distribution matching is employed to handle graphs with missing attribute~\cite{DBLP:journals/corr/abs-2011-01623}. In fact, augmentation strategies are applied in graph-based tasks for some time~\cite{DBLP:journals/tkde/FengHTC21,DBLP:conf/aaai/LuJ0S21,DBLP:journals/corr/abs-2111-06283}.
Chen \textit{et al.}~\cite{chen2020measuring} use optimized topology to alleviate the over-smoothing problem in graph representation learning. 
Graph Augmentation Learning (GAL) is a sort of graph learning approach that integrates a class of augmentation strategies, mechanisms, and models, aiming at overcoming the limitations and promoting the performance of graph learning models. 
Generally, GAL is designed to enhance the robustness of graph learning models with low-quality data by processing original data or modifying the graph learning models, or both.
GAL techniques can improve graph learning methods at all levels of range (i.e., nodes, edges, and graphs), thus achieving big progress in different graph learning applications.

Despite the extensive applications of GAL, the question of what scenario GAL techniques should be employed, and what GAL techniques are applicable, needs to be explored in depth. Unfortunately, there is no systematic, comprehensive, and experimentally validated guideline survey to refer to. 
Liu \textit{et al.}~\cite{DBLP:journals/corr/abs-2103-00111} and Xie \textit{et al.}~\cite{DBLP:journals/corr/abs-2102-10757} summarize the usage of graph self-supervised methods, which is one type of GAL technique, to enhance the performance of graph representation learning. Graph self-supervised method can provide a solution for few labels problem. But it is not the only choice in GAL. 
Survey papers of non-graph structured data in other domains (such as natural language processing) have provided a systematical view of using data augmentation to increase data diversity~\cite{DBLP:conf/acl/FengGWCVMH21,DBLP:conf/ijcai/Wen0YSGWX21}, but cases in graph representation learning are still missing.

To better understand the advantages and features of graph augmentation learning, we review the graph augmentation learning approach from two perspectives, i.e., GAL strategies and GAL application scenarios. 
Firstly, we analyze the main GAL strategies from the perspective of graph hierarchy, i.e., micro (node/edge), meso (subgraph), and macro (graph) levels. 
Then, we elaborate on how GAL works in different application scenarios. Specifically, we discuss data-specific, model-specific, and hybrid scenarios. In addition, we also implement experiments to show the outperformance of GAL methods in different application scenarios. A github repository of GAL source codes is also built~\footnote{https://github.com/yushuowiki/awesome-GAL}. Detailedly, our contributions are:
\begin{itemize}
	\item \textbf{Systematical Augmentation Strategy Introduction:} We systematically classify the strategies of existing GAL techniques from three structural levels (node/edge level, subgraph level, graph level) and further illustrate the differences between them through different hierarchies. To the best of our knowledge, this is the first review devoted to GAL.
	\item \textbf{Comprehensive GAL Application Description:} We give a comprehensive description of the existing application scenarios of GAL. In general, the different application scenarios can be considered as the three categories we have divided into, that are data-specific, model-specific, and hybrid scenarios.
	\item \textbf{Verifiable GAL Experiments Guidelines:} We perform experiments on GAL in various application scenarios and verify the effectiveness of GAL by comparing the results. We also analyze the experimental results and give further guidance for scholars to choose optimal GAL strategies.
\end{itemize}

The rest of this survey is organized as follows. 
In Section \ref{sec:gal_def}, we give the definition of graph augmentation learning. 
In Section \ref{sec:gal_stra}, we present the strategies used by GAL from three hierarchy levels, respectively. 
In Section \ref{sec:gal_in_rep}, we discuss how GAL methods perform in different graph learning scenarios. 
We run GAL experiments and give guidelines in Section \ref{sec:how2un}. 
In Section \ref{sec:open_issue}, we illustrate four open problems of GAL. 
We conclude the survey in Section \ref{sec:conclusion}. 

\section{What is Graph Augmentation Learning?}
\label{sec:gal_def}
As we have illustrated above, GAL can improve graph representation learning results and enhance robustness. Take Graph Convolutional Network (GCN)~\cite{DBLP:conf/iclr/KipfW17} as an example, the process of GAL is defined as Definition 1.

\textbf{Definition 1. (\emph{Graph Augmentation Learning):}} \label{def:GAL}

  For a given graph $\mathcal{G}=\{\mathcal{V},\mathcal{E},\mathbf{X}\}$, where $\mathcal{V}$ is a set containing $|\mathcal{V}|$ nodes, $\mathcal{E}$ is the set of edges showing the links between nodes, and $\mathbf{X}$ is the attribute matrix. 
  GAL technique $\mathrm{Aug}(\cdot)$ aims to learn a mapping function $\Phi:\Phi(\mathcal{V},\mathcal{E},\mathbf{X})\mapsto \mathbb{R}^{|\mathcal{V}|\times d}$ that projects graph nodes to $d$ dimension latent representation $\mathbf{Z}$, where $\Phi=\mathrm{Aug}(\Phi^{\prime})$, $\Phi^{\prime}$ is the GCN mapping function.


\begin{figure*}[h]
	\centering 
	  \includegraphics[scale=0.47]{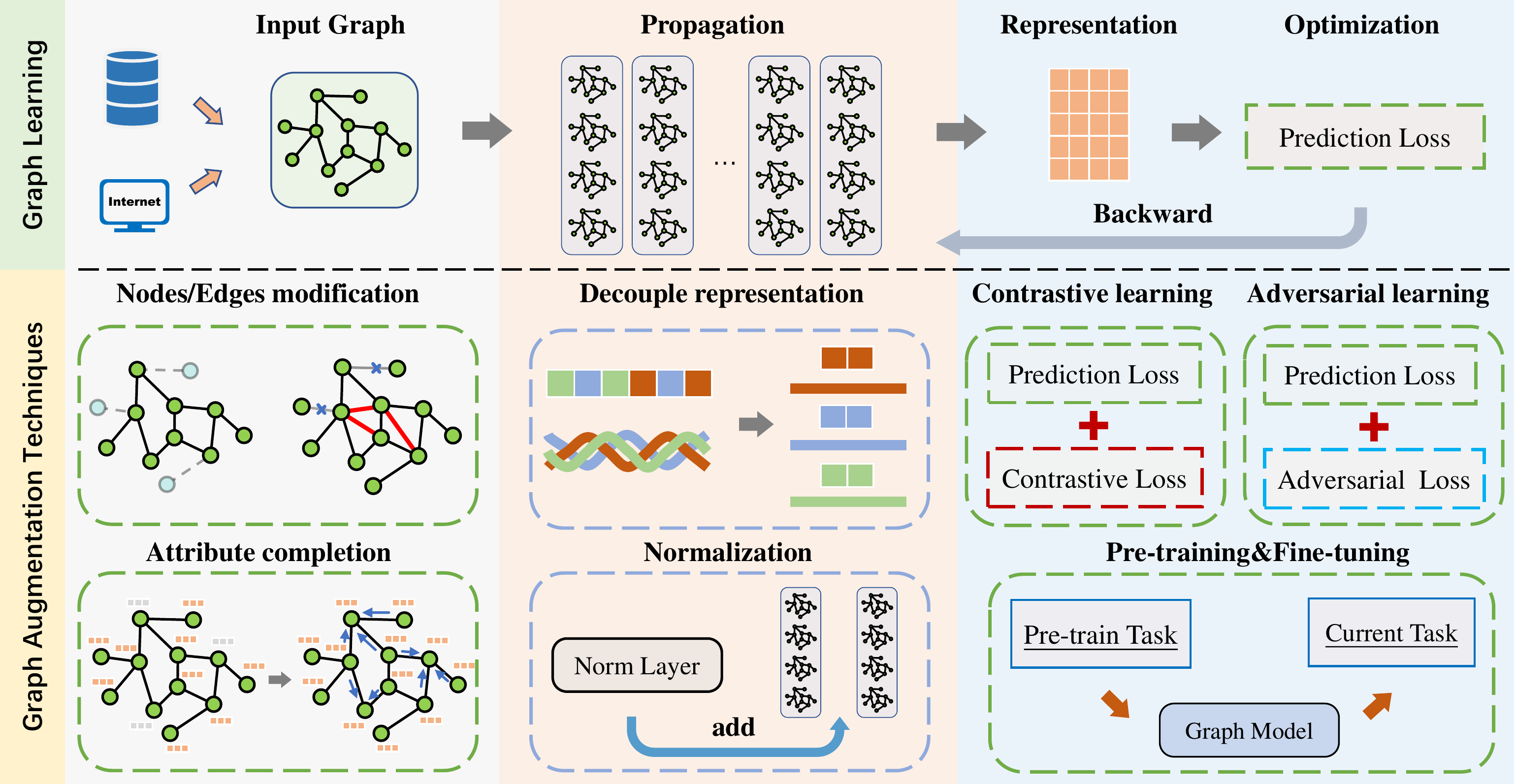}
	  \caption{An overall framework for GAL.}
  \label{fig:augment}
\end{figure*}

\section{An Overview of GAL Strategies}
\label{sec:gal_stra}

\begin{figure*}[h]
  \centering 
    \includegraphics[scale=0.47]{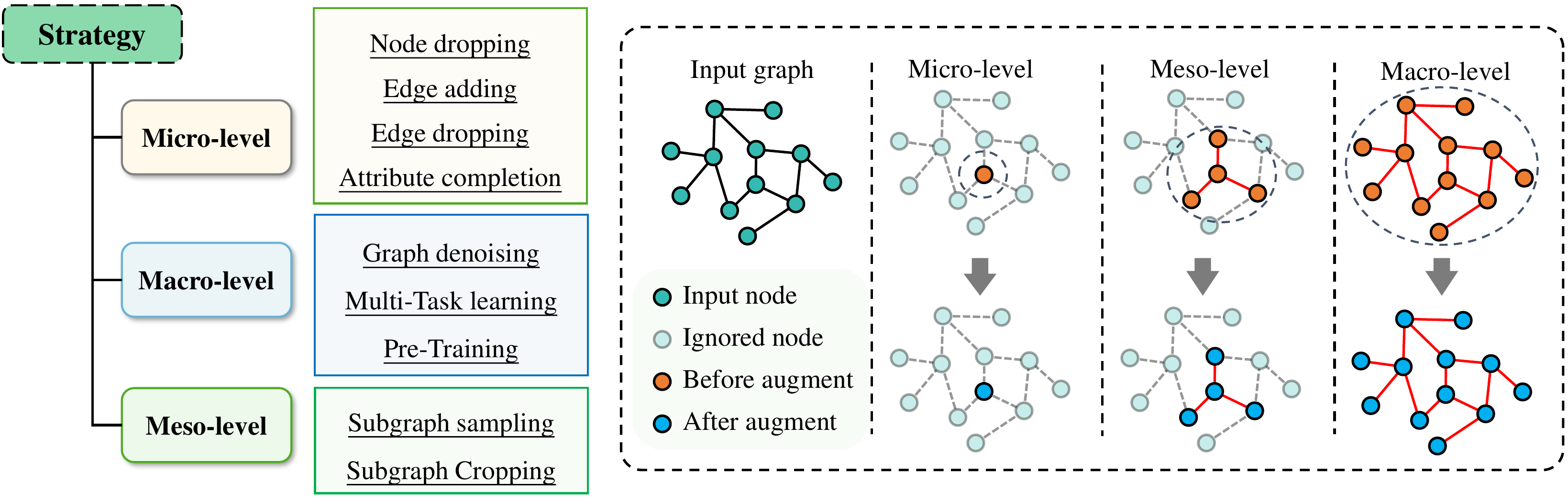}
    \caption{The overview of augmentation strategies in GAL.}
\label{fig:strategy}
\end{figure*}

In this section, we introduce three kinds of commonly-used augmentation strategies. Augmentation strategies for graph learning can be classified from three categories, i.e., \textbf{micro} (node/edge) level, \textbf{meso} (subgraph) level, and \textbf{macro} (graph) level, which is shown in Figure~\ref{fig:strategy}. The differences and characteristics of these augmentation approaches are respectively illustrated.

\subsection{Micro-level Augmentation Strategies}

Micro-level GAL techniques are dedicated to improving the performance of graph learning by focusing on changing subtleties of graphs (e.g., nodes, edges, and attributes). 
Specifically, DropGNN~\cite{DBLP:journals/corr/abs-2111-06283} improves the performance of Graph Neural Networks (GNNs) by running multiple experiments independently while dropping nodes randomly. 
DropEdge~\cite{DBLP:conf/iclr/RongHXH20} randomly removes graph edges in message passing mechanism to alleviate over-smoothing.  
In addition, Dropout~\cite{DBLP:journals/jmlr/SrivastavaHKSS14} prevents overfitting through discarding a fixed portion of attributes (or neurons), which has become a widely used method in the field of deep learning. 
Micro-level GAL techniques have been widely used because the augmentation mechanism is straightforward and easy to realize. Compared to meso and macro level augmentation strategies, micro-level ones are generally in lower computational costs. Meanwhile, implementing micro-level strategies can also achieve satisfying performance.
Since such kind of augmentation strategy function on micro-level is not considering graph structures, it can be employed in abundant graphs such as hypergraphs and heterogenous graphs. Accordingly, it has broader applications than the other two kinds, and can effectively improve the robustness as well as generalization ability of models.

\subsection{Meso-level Augmentation Strategies}

Meso-level GAL techniques, compared with the micro-level one, generally employ path or subgraph information (of target nodes/edges) to enhance graph representation learning. 
There are some studies using efficient subgraph sampling and computing methods to accelerate meso-level augmentation strategies, thus reducing computational and spatial costs to be affordable. 
Some studies sample neighbors for each node in the graph, generate low-order subgraph structures, and then execute message passing operations in these subgraphs~\cite{zhang2021nested,DBLP:conf/aaai/YouGYL21}.  
Similarly, Yu \textit{et al.}~\cite{DBLP:conf/cikm/Yu0XCL20} and Xu \textit{et al.}~\cite{DBLP:conf/jcdl/XuYSRLP020} use motifs to effectively capture the higher-order relationship patterns in complex networks. 
GraphCrop~\cite{DBLP:journals/corr/abs-2009-10564}, by pruning the subgraph, removes the massive noise in graphs and achieves better graph classification performance. 
Compared with micro-level, meso-level GAL techniques pay attention to the wider range of information around nodes in the graph. Subgraph information differs significantly in different kinds of networks. The semantic information of triangle structure is obviously different in social network and traffic network. Therefore, meso-level strategies can extract personalized information in a certain kind of network compared to micro-level ones. On the other hand, the universality is weaker than that of micro-level strategies. As a consequence, this kind of strategy is more popular in recommendation systems and neural language processing.

\subsection{Macro-level Augmentation Strategies}
 
Macro-level GAL strategies target to improve graph learning methods from a global view. Concretely, macro-level strategy focuses on graph-level modification or other global operations in graph learning. 
Wang \textit{et al.}~\cite{DBLP:conf/kdd/0017ZB0SP20} use node attribute to construct a K-Nearest Neighbor (KNN) graph, thereby enhancing the model's learning ability without modifying the original graph topology in the graph learning process.
At the same time, some studies use joint optimization of multiple highly related tasks to estimate the missing information in a single task~\cite{DBLP:conf/www/JinHL021,DBLP:conf/kdd/Jin0LTWT20}, others use pre-training~\cite{DBLP:conf/aaai/LuJ0S21} to accelerate the learning process. 
In addition, self-supervised methods~\cite{sun2020multi} can be used as another competitive macro-level augmentation strategy.

Moreover, multiple strategies can also be employed together to achieve better performance~\cite{DBLP:journals/corr/abs-2011-01623,DBLP:conf/icml/YouCWS20}, because they can integrate different levels of useful information. The common advantages of GAL strategies are summarized as follows.
\begin{itemize}
  \item \textbf{Auxiliary:} GAL strategies are not newly proposed graph learning methods. Instead, augmentation learning aims to enhance the performance and robustness of graph learning methods.
  \item \textbf{Pluggability:} GAL strategies rarely modify the original graph learning model architecture and meanwhile achieve meaningful improvements. For a given augmented solution, when removing the augmentation part, the model will also function normally. The ability of the original model to complete downstream tasks will not be totally damaged.
  \item \textbf{Generality:} The pluggability of GAL strategies determines generality to some extent. Similar graph learning models (such as GCN and GAT) can share the same GAL strategies. As a result, the improvements in augmentation are also similar.
 \end{itemize}
 
\section{Augmentation in Graph Learning}
\label{sec:gal_in_rep}
In this section, we discuss how the GAL outperforms original graph learning methods. Specifically, three different situations are respectively analyzed: (1) GAL for low-quality data, (2) GAL for model limitation, and (3) GAL for both.

\subsection{GAL for Low-quality Data}
Real world data are generally low quality because of various reasons such as data collection loss and privacy protection policies. These data limits the performance of graph learning at the first beginning and generally leads to poor learning results. The majority of existing graph learning algorithms presume that the data is almost anomaly-free and of high quality. However, lost and erroneous overlays are unavoidable during data collection and transmission. Moreover, the graph data naturally contains different degrees of erroneous or noisy information. Once the impact of low-quality data on graph learning tasks is ignored, the optimization process of learning algorithms can easily be hindered or even misled, leading to sub-optimal or even worse results in complex real-world scenarios. 

Therefore, GAL techniques need to handle low-quality data while maintaining efficiency and robustness. At present, there have been plenty of studies in Computer Vision (CV) domain to investigate how to overcome low-quality data limitations. But graph data are very different from images, thus few augmentation methods can be employed. Herein, we discuss GAL methods under three main types of data limitations: (1) low-quality node/edge attributes, (2) low-quality graph structures, and (3) few-labeled data. 

\subsubsection{Low-quality Node/edge Attributes.} 
 
Attribute is one of the most important factors in graph learning. Attributes can directly reflect the basic information as well as key features of nodes/edges in the graph. Intuitively, the attribute information in the graph data from the web is sometimes partially corrupted or completely damaged, leading to data low-quality problems. Low-quality attributes often lead to bad performance in almost all kinds of graph learning methods. In order to mitigate the negative impact of low-quality data, Chen \textit{et al.}~\cite{DBLP:journals/corr/abs-2011-01623} develop a novel GNN using a distribution matching technique for attribute-missing graphs. Experimental results indicate the effectiveness of the proposed method in link prediction and attribute completion tasks. HGNN-AC~\cite{DBLP:conf/www/JinHL021} use graph attention mechanism to complete the missing attributes of nodes in heterogeneous graphs, avoiding using previous hand-crafted ways to solve this problem.

\subsubsection{Low-quality Graph Structures.} 

A slight change in graph structure can cause significant differences, due to the fact that the data samples are relational to each other. Compared to node attributes, low-quality graph structures pose greater challenges for graph learning. This is because generally connected nodes are regarded to have similar attributes, but not vice versa. Nodes with similar attributes might connect with each other in the further but this is not absolute. Therefore, structural information are usually employed to handle low-quality attribute problem, but attributes may be not that useful in dealing with low-quality structures. It is necessary to denoise the graph structure to extract useful and helpful information correctly. 

Nevertheless, modifying or even reconstructing the topology becomes another competitive choice for GAL methods to the graph tasks. Luo \textit{et al.}~\cite{DBLP:conf/wsdm/LuoCYZNCZ21} introduce a denoising network learning to drop edges that are irrelevant to the downstream tasks, thus improving the robustness and generalization capability in GNNs. Zhao \textit{et al.}~\cite{DBLP:conf/aaai/0003LNW0S21} apply an edge predictor to modify input graphs for better node classification performance. Wang \textit{et al.}~\cite{DBLP:journals/corr/abs-2009-10564} obtain various subgraphs by cropping operation and generate information-rich augmented graphs to expand the training set. In short, by adding and deleting edges, the structural noise of graph data can be eliminated to a proper extent. In this way, graph learning models can achieve better performance with high robustness at the same time.

\subsubsection{Few-labeled Data.}
Though a huge amount of data is easily available in the era of big data, sometimes there is only a small percentage is usable. Since the data labeling cost is too high, labeled datasets are often in limited size. Especially in some special research areas, such as biochemistry and medicine, there is only a small amount of available labeled data. The scarcity of labeled data makes graph learning models inevitable prone to overfitting and poor robustness during the training process. Many approaches have been proposed to overcome the limitations caused by insufficient data, i.e., few-labeled data. In the following, we describe how to use GAL to deal with insufficient data from both supervised and unsupervised perspectives. Specifically, we summarize the existing methods in Table~\ref{tab:FewDataQuantity}.

\begin{table}[]
  \centering
	\caption{GAL for few-labeled data.}
  \begin{tabular}{cll}
  \toprule
  \multicolumn{1}{l}{}                               & \textbf{Types}                 & \textbf{Methods}                      \\ \midrule
  \multicolumn{1}{c|}{\multirow{2}{*}{\textbf{Supervised}}}   & \multicolumn{1}{l|}{Pretrain}  & \cite{hu2020strategies},~\cite{hu2020gpt}          \\
  \multicolumn{1}{c|}{}                              & \multicolumn{1}{l|}{Resist}    & \cite{gao2021training},~\cite{wang2020nodeaug},~\cite{DBLP:conf/icml/ZhengZCSNYC020}        \\ 
  \multicolumn{1}{c|}{\multirow{2}{*}{\textbf{Unsupervised}}} & \multicolumn{1}{l|}{Predefine} & \cite{you2020graph},~\cite{zeng2021contrastive},~\cite{kong2020flag} \\
  \multicolumn{1}{c|}{}                              & \multicolumn{1}{l|}{Adaptive}  & \cite{zhu2021graph},~\cite{you2021graph},~\cite{suresh2021adversarial}  \\ \bottomrule
  \end{tabular}
  \label{tab:FewDataQuantity}%
\end{table}

\textit{Supervised Methods.} 
In supervised learning settings, when the data amount in the training set is not enough, graph learning methods cannot learn sufficiently in the training process. Moreover, models tend to overfit, leading to poor performance on the test dataset. For few-labeled data, expanding the dataset and enhancing the quality of available data are popular choices. GAL can be used to augment graph data, thus solving the defects caused by data sparsity. ``Pre-training plus fine-tuning" is an excellent paradigm in GAL to solve few-labeled data problems. In this process, the model is first trained on abundant data from other relevant domains and then fine-tuned in the target downstream task using only a small amount of data. Hu \textit{et al.}~\cite{hu2020strategies} combine node-level and graph-level pre-training and achieve success in molecular property prediction and protein function prediction. GPT-GNN~\cite{hu2020gpt} employs self-supervised graph generation to pre-train GNN and applies it in recommender systems. In addition, improving the ability to resist overfitting in model training has become another popular solution choice. Injecting perturbations and noise into the graph, such as adding edges and removing nodes, can increase the diversity of the training data and thus contribute to improving the robustness of the model. TADropEdge~\cite{gao2021training} drops graph edges adaptively based on their weights by graph spectrum-based computation, and improves generalization performance. Wang \textit{et al.}~\cite{wang2020nodeaug} propose three GAL techniques on the graph data by adjusting both the node attributes and the graph structure, thus yielding significant gains for GCN models. There are also exist methods that utilize adversarial training to accomplish data expansion and can be used in large-scale datasets~\cite{kong2020flag}.

\textit{Unsupervised methods.} 
Among the unsupervised learning methods, contrast learning has been successfully prevalent in CV, Natural Language Processing (NLP), and other fields. Graph Contrast Learning (GCL), one of the important GAL techniques, has also received increasing attention. The major idea in GCL is to achieve augmentation by contrasting representations extracted from the same graph in different views. Predefined augmentation strategies are popular in GCL and can represent solutions that are designed in advance. It covers different hierarchical levels (nodes, subgraphs, and graphs) and often uses multiple strategies together, such as falling edges, feature masks, subgraph sampling, and graph diffusion~\cite{you2020graph,zeng2021contrastive}. However, how to choose and design an appropriate GAL becomes a question worth discussing. Moreover, the predefined GAL strategies used in most existing solutions lack flexibility. Therefore, dynamic learning of optimal augmentation strategies during the training process is another hot research issue~\cite{suresh2021adversarial}. Therefore, adaptive GAL methods~\cite{zhu2021graph,you2021graph}, e.g., automatically selecting the optimal combination of augmentation strategies for the currently running task, are becoming more and more popular. 

\subsection{GAL for Model Limitation}

GNNs are successful paradigms of deep learning on graph data, with excellent performance in social networks, traffic networks, etc. GNNs have become prevalent approaches in graph learning. Despite their excellent learning capability, the limitations of GNN models have been emerging. For example, as the layer number increases, the performance of many powerful GNN models does not get better or even degrades in node classification tasks due to over-smoothing. In addition, the message passing mechanism in GNNs borrows from the 1-dimensional Weisfeiler-Lehman (1-WL) algorithm~\cite{leman1968reduction}, which absolutely gives GNNs excellent capabilities, but also makes them unable to distinguish some special isomorphic graphs. Therefore, GNNs expressive power is accordingly limited in many situations. In this section, we discuss how to solve the above problems through GAL methods.

\subsubsection{Over-smoothing}
As the depth of the model increases, the representations of all nodes in the graph gradually fused and eventually become indistinguishable with iterative message passing. This is the so-called over-smoothing phenomenon, and such a phenomenon generally leads to failure in the task~\cite{DBLP:conf/aaai/LiHW18}. To cope with over-smoothing, adding noise to the data or giving regularization terms to the optimization process are effective means.

\begin{table}[htbp]
  \centering
	\caption{GAL for over-smoothing.}
  \begin{tabular}{cll}
  \toprule
                                                           & \textbf{Types}                      & \textbf{Methods}            \\ \midrule
  \multicolumn{1}{c|}{\multirow{4}{*}{\textbf{Locality}}}  & \multicolumn{1}{l|}{Perturbation}   & \cite{liu2021graph},~\cite{lu2021skipnode}         \\
  \multicolumn{1}{c|}{}                                    & \multicolumn{1}{l|}{Propagation}    & \cite{chen2020simple},~\cite{yang2021attributes},~\cite{hu2020going} \\
  \multicolumn{1}{c|}{}                                    & \multicolumn{1}{l|}{Transformation} & \cite{liu2020towards},~\cite{min2020scattering}          \\
  \multicolumn{1}{c|}{}                                    & \multicolumn{1}{l|}{Subgraph-based} & \cite{zeng2020deep},~\cite{zhou2020towards}          \\
  \multicolumn{1}{c|}{\textbf{Globality}}                  & \multicolumn{1}{l|}{Regularization} & \cite{zheng2021tackling},~\cite{chen2020measuring},~\cite{DBLP:conf/iclr/VelickovicFHLBH19}    \\ \bottomrule
  \end{tabular}
  \label{tab:OverSmoothing}%
\end{table}

\textit{Locality.} From a local perspective, we classify the GAL methods into perturbation, propagation, transformation, and subgraph-based methods. To facilitate understanding, we show how propagation and transformation work in graph convolution layer by follows:

\begin{equation}
  \begin{aligned}
    &a_{i}^{(l)}=\textbf { Propagation }\left(\left\{h_{i}^{(l-1)}\right\}\cup\left\{h_{j}^{(l-1)} \mid j \in \mathcal{N}_{i}\right\}\right) \\
    &h_{i}^{(l)}=\textbf { Transformation }\left(h_{i}^{(l)}\right),
    \end{aligned}
    \label{pro_tran}
\end{equation}
where $h_i^{(l)}$ represents the feature of node $i$ in $l$-th layer, and $\mathcal{N}_{i}$ is neighbor node set from node $i$. Next we introduce these types of above-mentioned methods.

\textbf{Perturbation}, as the name suggests, this sort of method randomly changes the nodes/edges contained in the original graph during the training process, or masks some of the nodes from computation. For instance, Liu \textit{et al.}~\cite{liu2021graph} dynamically add and remove edges based on the graph motif structure to preserve the graph entropy as much as possible. SkipNode~\cite{lu2021skipnode}, on the other hand, lets some nodes skip the graph convolution operation in a certain probability, preserving the features for the next use, thus mitigating the gradient vanishing and weight over-decaying issues. 
\textbf{Propagation} is an important component of convolution on a graph, i.e., the way nodes aggregate neighbors, typically includes direct neighbors or multi-hop neighbors. Some special propagation methods are carefully designed to prevent over-smoothing. GCNII~\cite{chen2020simple} aggregates the initial and the previous layer representation to the current layer by adding a residual connection in training. Yang \textit{et al.}~\cite{yang2021attributes} then propose graph conjugate convolution based on GCNII. 
\textbf{Transformation} represents the way how features are transformed in graph convolution and plays a key role in graph learning. Some methods try to solve over-smoothing by improving the graph convolution computation formula. Liu \textit{et al.}~\cite{liu2020towards} successfully increase the model depth by decoupling graph convolution while maintaining performance and avoiding over-smoothing. Scattering GCN~\cite{min2020scattering} integrates geometric scattering transforms and residual convolutions to augment conventional GCNs.
Subgraph-based method is a type of approach that improves the graph representation learning capability by considering \textbf{subgraph information}. For example, SHADOW-GNN~\cite{zeng2020deep} samples subgraphs centered on each node in the whole graph and then builds an L-layer GNN operating on subgraph instead of the whole graph. Zhou \textit{et al.}~\cite{zhou2020towards} introduce a differentiable group normalization to GNNs, which not only normalizes nodes within the sample group independently but also separates node distributions among different groups to alleviate over-smoothing.

\textit{Globality.} 
Regularization, which is widely used in machine learning to prevent overfitting, can also mitigate the over-smoothing problem from a global perspective. By adding a regularization term to the loss function in a deep model, the node representations' fast convergence can be prevented. For example, TGCL~\cite{zheng2021tackling} adds a graph contrastive layer guided by topology before the output layer of the GNN model, reducing the negative impact of remote nodes with similar topology features. In addition, Chen \textit{et al.}~\cite{chen2020measuring} point out that over-smoothing is caused by an excessive mixture of information and noise, and use a regularization term to increase the received information and decrease the received noise. There are also some studies using other types of GAL strategies. The cause of over-smoothing is regarded as the entanglement of propagation and transformation~\cite{liu2020towards}. By decoupling these two operations, the model can avoid over-smoothing. In general, most of these approaches mentioned above can be unified as follows:
\begin{equation}
	\begin{aligned}
		\mathbf{H}^{l+1} &= \sigma\left(\mathcal{T}_1(\mathbf{A})\mathcal{T}_2(\mathbf{H}^l)\mathbf{W}^l\right)\\
		\mathcal{L}_{total} &= \mathcal{L}_{rep} + \mathcal{L}_{reg}, 
	\end{aligned}
\end{equation}
where $\mathbf{A}$ refers to the adjacency matrix, $\mathbf{H}^l$ and $\mathbf{W}^l$ are the weight and hidden representation of layer $l$, respectively. $\mathcal{T}_1,\mathcal{T}_2\in\mathcal{T}$ represent specifical augmentation strategies like drop node, drop edge or mask feature, etc., $\mathcal{L}_{rep}$ denote the representation learning loss, and $\mathcal{L}_{reg}$ is a regularization term to relieve over-smoothing.

\begin{figure}[h]
		\centering 
		\includegraphics[scale=0.35]{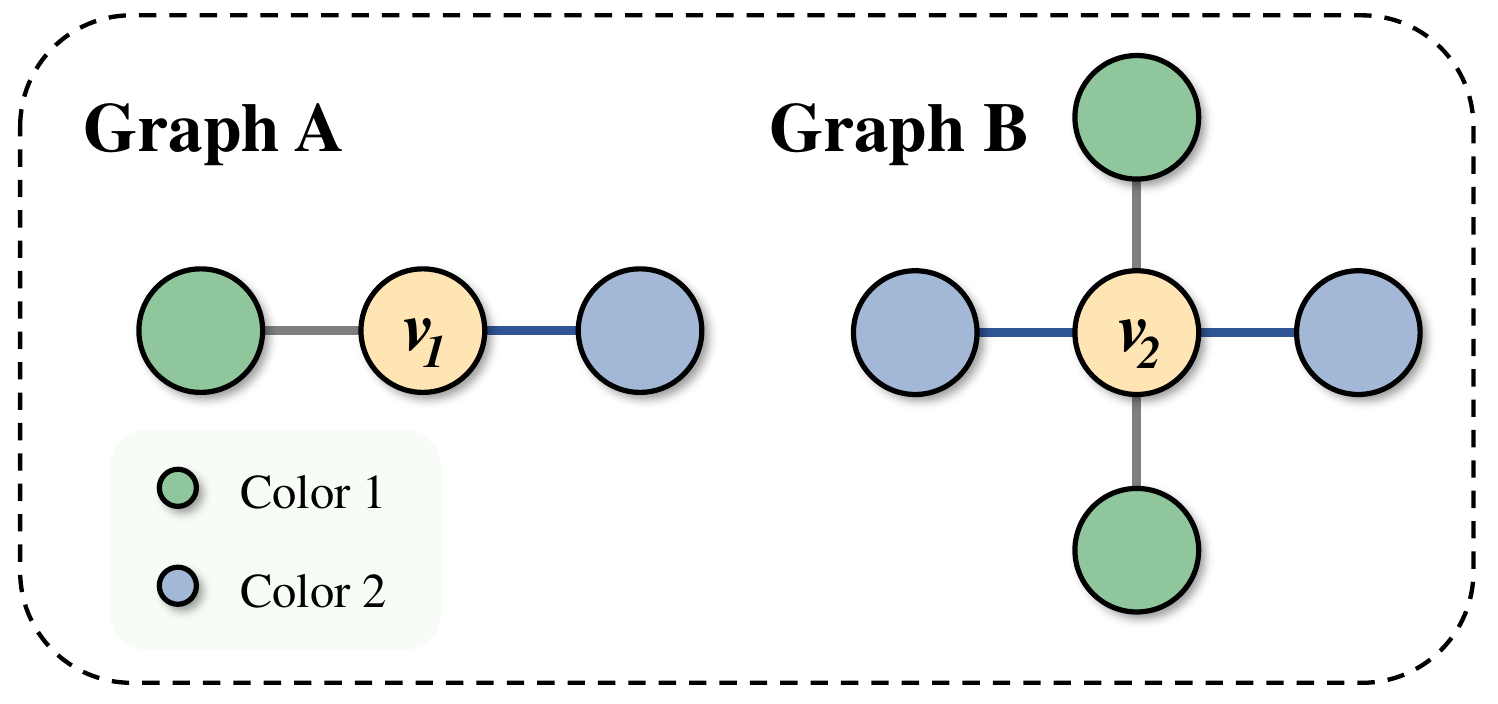}
		\caption{Two networks that traditional GNNs fail to distinguish.}
	\label{fig:isomorphism_graph}
\end{figure}

\subsubsection{1-WL}
GNN models have limitations in representation ability. Many pieces of research show that GNNs are neural variants of 1-WL, the algorithm used to test whether two graphs are isomorphic. They also show that GNNs do not exceed the expressive power of 1-WL~\cite{morris2019weisfeiler,xu2018powerful}. For instance, WL-based GNNs, such as GraphSAGE~\cite{DBLP:conf/nips/HamiltonYL17}, JKNet~\cite{DBLP:conf/nips/DuvenaudMABHAA15}, and GAT~\cite{DBLP:conf/iclr/VelickovicCCRLB18}, cannot distinguish the two graphs in Figure~\ref{fig:isomorphism_graph}. But the difference can be identified by 1-WL. In this section, we discuss how to improve models through GAL and obtain GNNs with comparable or even better capabilities than 1-WL.

Some GNN algorithms fail to distinguish the two graphs because they use non-injective readout functions (e.g., mean, max, or sum) to transform node information to graph representation, and thus they cannot adapt to complex situations. GAL can be an outstanding choice to tackle this challenge. The approaches of such GALs vary, but most of them have one thing in common, i.e., they improve the way the aggregation and transformation operation in GNNs. In the following, we present how to address the limitations of GNNs' representation ability through GAL techniques.

One feasible attempt is to design an excellent message passing scheme to obtain a GNN model with better performance and to catch up (or outperform) 1-WL in distinguishing non-isomorphic graphs. You \textit{et al.}~\cite{DBLP:conf/aaai/YouGYL21} propose ID-GNN, which is more powerful than 1-WL test. ID-GNN assigns a unique identity to each node, which is used to distinguish the node from other nodes in its neighbors, and the node's identity is taken into account to update the node's embedding during the GNN's message passing process. Maron \textit{et al.}~\cite{maron2019provably} design a simple GNN model that utilizes a multilayer perceptron (MLP) and matrix multiplication, which has a provable 3-WL expressive power while maintaining scalability. 

Considering that the GNN model is a variant of 1-WL, one way to improve it is to design a GNN based on $k$ ($k>1$) dimensional Weisfeiler-Lehman ($k$-WL), which can be implemented by subgraph-based or local structural augmentation technology. Some studies claim that the reason behind such limitation of GNNs lies in the inability to adequately capture graph topology information. It can be resolved by subgraph sampling and other GAL strategies. These studies usually introduce graph structure information during feature aggregation iteration by comparing WL algorithms and GNNs through theoretical analysis. In particular, Zhang and Li~\cite{zhang2021nested} propose NestedGNN, which embeds nodes by encoding a rooted subgraph instead of a rooted subtree. NestedGNN is more powerful than message passing GNNs and 1-WL. Inspired by $k$-WL, $k$-GNN~\cite{morris2019weisfeiler} considers multi-scale higher-order graph structures and strictly outperforms GNNs in the ability to distinguish non-isomorphic graphs/subgraphs. 

\begin{figure*}[]
    \centering
  \subfloat[Node classification
  ]{
    \label{fig:nc_box}
    \includegraphics[width=2.1in]{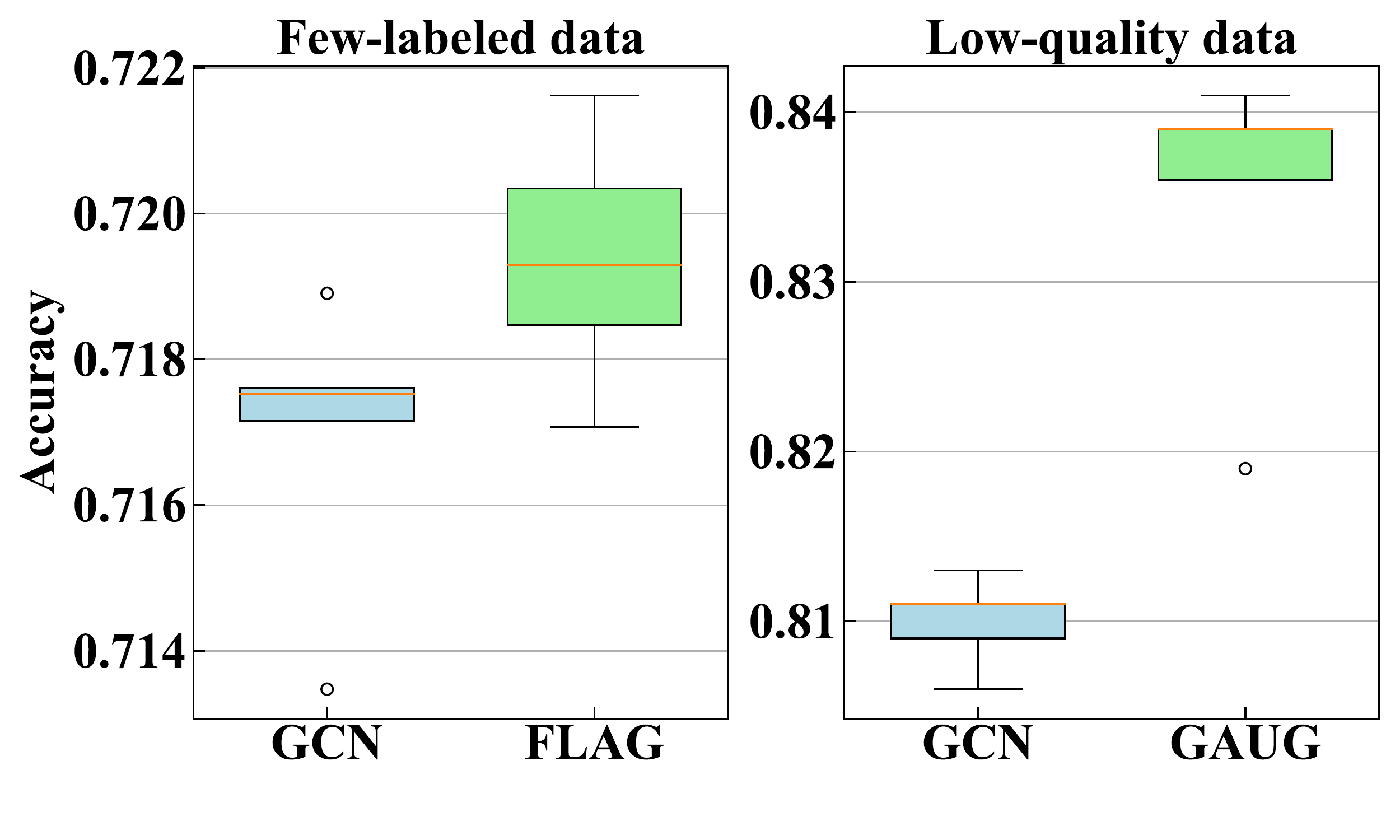}
    }
  \subfloat[Graph classification]{
    \label{fig:gn_box}
    \includegraphics[width=2.1in]{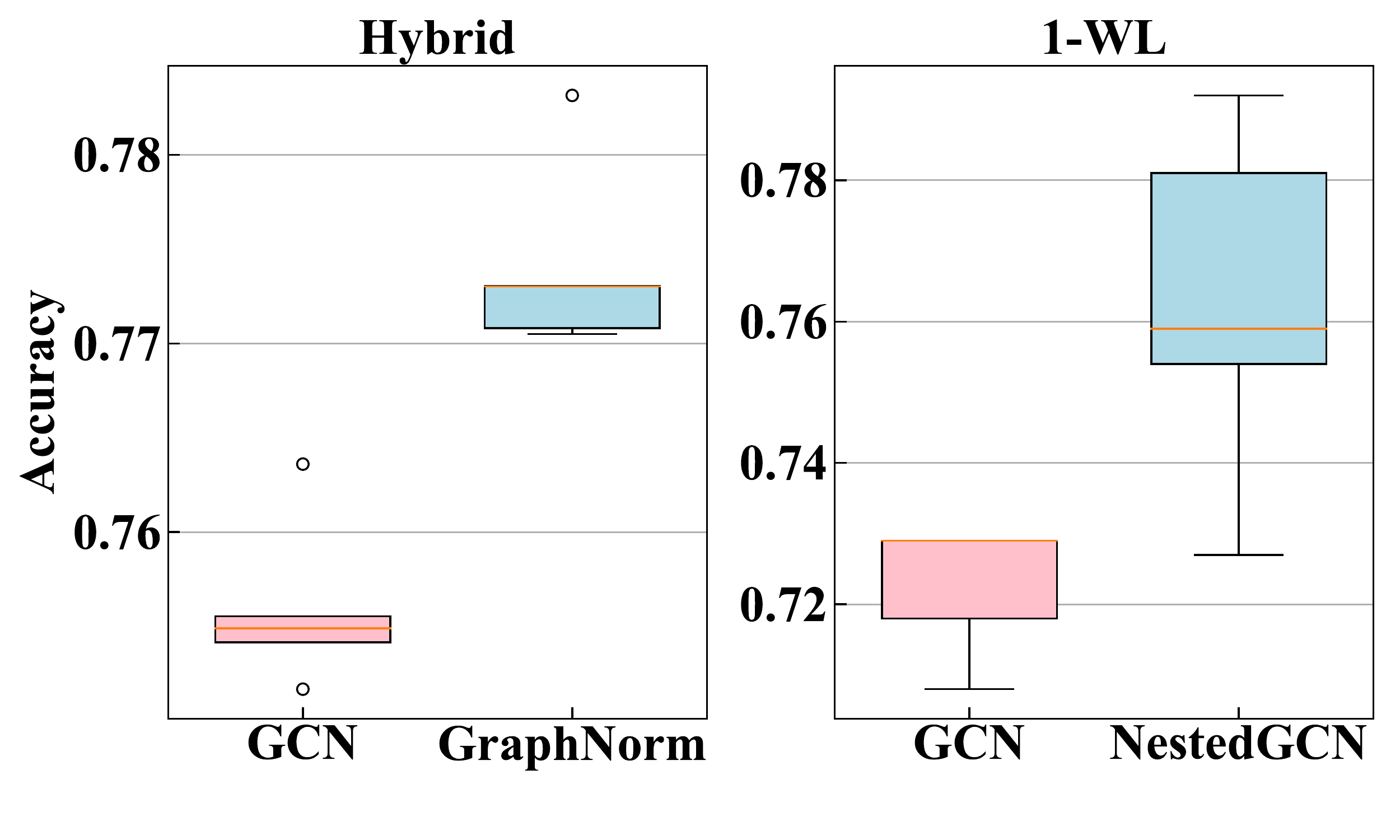}
    }
  \subfloat[Over-smoothing]{
    \label{fig:ovsms}
    \includegraphics[width=2.1in]{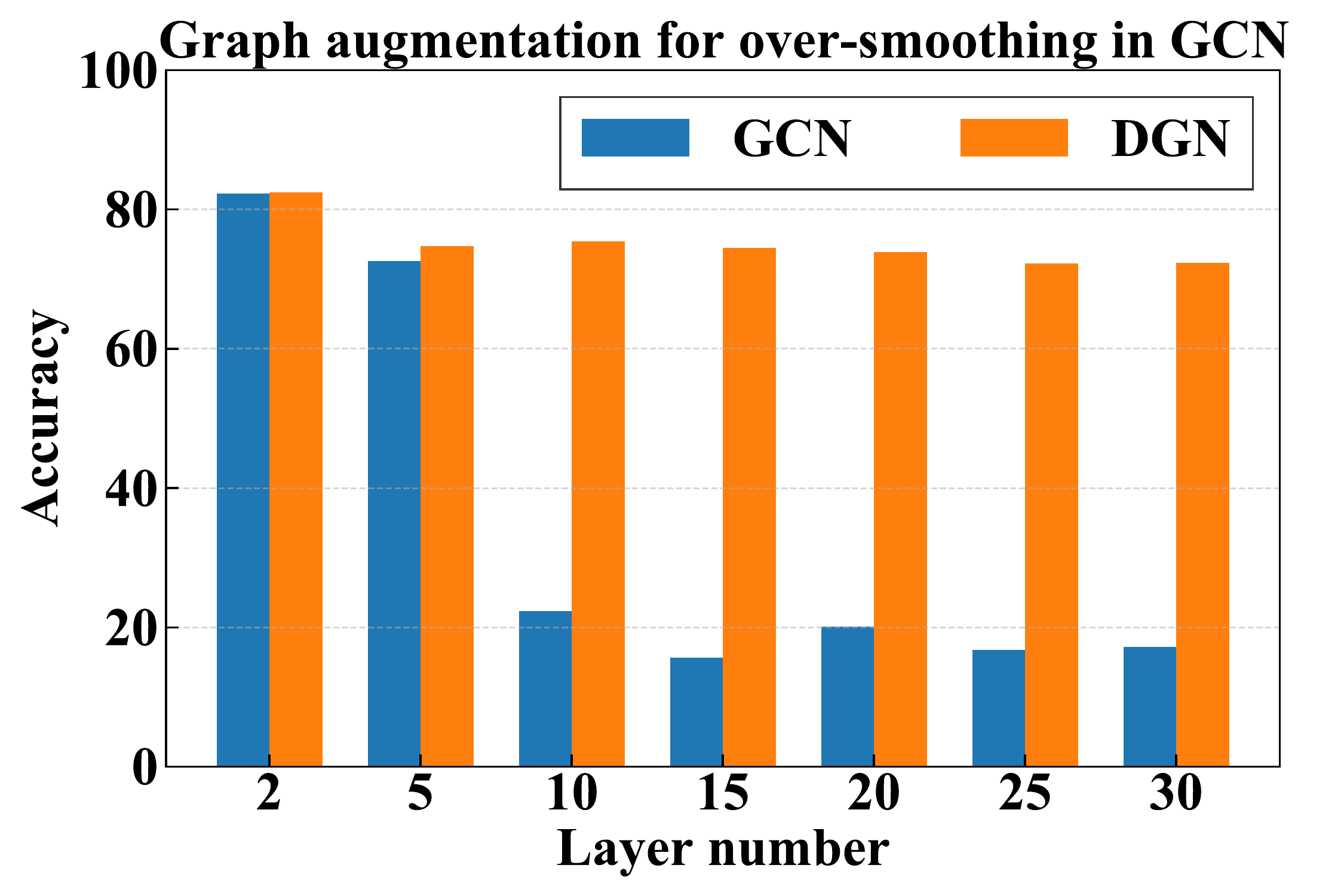}
    }
	\caption{Graph augmentation learning results in different tasks.}
  \label{fig:experiment}
\end{figure*}

\subsection{GAL for Hybrid Scenario}
In most cases, low-quality data and model limitation both exist in graph learning tasks. But most current studies are devoted to solving the problem from one perspective. Those studies directed at solving the graph learning model problem are specified in terms of expanding node perceptual field, improving expressiveness, and preventing over-smoothing. However, these works generally consider graph data are perfect, which is impracticable in real scenarios, like bot users and invalid users in social networks. Moreover, there may be insufficient supervised labels in the data, such as drug network data and biomolecular network data. Therefore, in real-world scenarios, the challenges of data and models appear in a mixed manner. 

Designing solutions for only one class of problems while ignoring the other one is clearly lacking in generalization, and will eventually lead to suboptimal results. Thus, we need solutions that can solve the model problem and at the same time can accommodate the shortcomings of low-quality graph data. Numerous hybrid scenarios require augmentation techniques, but only a few existing works study such scenarios simultaneously.

In this regard, Jin \textit{et al.}~\cite{DBLP:conf/aaai/0002C0S21} analyze the limitations of Laplace operator in GCN and point out that the operator is not robust to the noise links existing in graph data. Therefore, they design a new operator and integrate it into the existing GCN backbone, to improve the noise tolerance of data and the performance of the model. Normalization technology can accelerate the optimization process and has been widely used in deep learning, especially in the fields of CV and NLP. For graph data, GraphNorm~\cite{DBLP:conf/icml/CaiLXHL021} successfully adapts to noise and high variance through processing graph batch in tasks for complex data environments. The generalization ability of the model is also enhanced.

\section{How to Understand GAL?}
\label{sec:how2un}
In this section, we further verify the performance of GAL techniques and then give sound advice on choosing GAL strategies. We have presented the strategies of GAL from three perspectives in Section \ref{sec:gal_stra} and illustrated that GAL can solve different problems for graph learning in Section \ref{sec:gal_in_rep}. It is obvious that GAL, as an efficient tool, can remedy the shortcomings of existing graph learning methods in various scenarios. However, there still remain many questions. How to choose GAL strategies in certain applications? How much degree GAL can improve the performance? 

Since GAL can overcome the limitations existing in the original graph learning method, More explicitly, GAL can bring a direct experimental performance gain for the task. Therefore, we run experiments with the graph learning approach in each of the scenarios mentioned in this paper and quantitatively evaluate the gains brought by the GAL to the original methods. Afterward, we analyze the experiment results and give guiding suggestions for future researchers on graph augmentation techniques in various scenarios.

For this purpose, we select datasets for performing GAL experiments for mentioned scenarios in Section \ref{sec:gal_in_rep}. These challenging cases include few-labeled data, low-quality data, over-smoothing, 1-WL limitation, and hybrid scenario, respectively. 
The dataset used in the experiments are Cora~\cite{DBLP:conf/iclr/KipfW17}, ogbn-Arxiv~\cite{DBLP:conf/nips/HuFZDRLCL20}, ogbg-MOLHIV~\cite{DBLP:conf/nips/HuFZDRLCL20}, and MUTAG~\cite{DBLP:journals/corr/abs-2007-08663} as shown in Table \ref{tab:data}.


\begin{table}[]
	\centering
  \caption{Statistic information of datasets used in the experiments.}
  \begin{tabular}{cccc}
  \toprule
  \textbf{Datasets}       & \textbf{\#Graphs} & \textbf{\#nodes}      & \textbf{\#edges}    \\ \midrule
  \multicolumn{1}{c|}{ogbn-Arxiv}  & 1        & 169,343      & 1,166,243  \\
  \multicolumn{1}{c|}{Cora}        & 1        & 2,708        & 5,429      \\ \bottomrule
  \toprule
  \textbf{Datasets}       & \textbf{\#Graphs} & \textbf{Avg. \#nodes}      & \textbf{Avg. \#edges}    \\ \midrule
  \multicolumn{1}{c|}{ogbg-MOLHIV} & 41,127   & 25.5    & 27.5  \\
  \multicolumn{1}{c|}{MUTAG}       & 188      & 17.93   & 19.79 \\ \bottomrule
  \end{tabular}
  \label{tab:data}
\end{table}


In all the experiments, we choose GCN as the original solution, which is widely used in the study of graph learning. After augmentation, we get augmented GCN methods and then use them to run experiments. The results are compared with those output by vanilla GCN. Firstly, in node classification (NC) task (with few-labeled data) we choose the method proposed by Kong \textit{et al.}~\cite{kong2020flag}. For the limitation of low-quality data, GAUG~\cite{DBLP:conf/aaai/0003LNW0S21} is implemented. In the graph classification (GC) task, we use NestedGCN~\cite{zhang2021nested} to augment the GCN in order to cope with the 1-WL capability limitation. For the hybrid case, we apply GraphNorm proposed by Cai \textit{et al.}~\cite{DBLP:conf/icml/CaiLXHL021}. In addition, for the above four methods, we run them five times using different random seeds. Finally, we use DGN~\cite{zhou2020towards} to handle the over-smoothing problem in node representation.

Figure \ref{fig:experiment} shows the results of experiments. FLAG~\footnote{https://github.com/devnkong/FLAG} augments the features of graph nodes by introducing adversarial perturbations in training. It improves the GNN performance while preventing overfitting due to few-labeled data. 
For the low-quality data problem, GAUG~\footnote{https://github.com/zhao-tong/GAug} proposes an edge predictor that achieves performance improvement by denoising graph edges, i.e., removing noisy edges and adding edges that potentially missing from the original graph. 
GraphNorm~\footnote{https://github.com/lsj2408/GraphNorm} proposes a learnable shift to normalize the node representation in the graph, handles the noise in the graph that cannot be coped with BatchNorm. 
Such an improvement also avoids the expressiveness degradation caused by the shift operation brought in InstanceNorm and allows the algorithm to converge faster. 
NestedGCN~\footnote{https://github.com/muhanzhang/NestedGNN} first extracts subgraphs for each node, then learns the subgraph representation, and finally, through a message passing mechanism, gets the graph representation. Such an approach makes it beyond the 1-WL power. 
DGN~\footnote{https://github.com/Kaixiong-Zhou/DGN} takes into account the distribution of different communities in the graph and takes different normalization operations for various communities, thus resist to over-smoothing. 
Overall, we can see that the task metrics of the backbone are improved after using GAL techniques, which reflects the efficiency of graph augmentation learning. 


The above experimental results reflect that GAL can enhance GNN in different tasks. Here we give guiding suggestions for using GAL technology based on the experimental results and related work. 
\begin{itemize}
  \item \textbf{Few-labeled data.} Training techniques can be employed to compensate for the lack of labeled data, e.g., pre-training plus fine-tuning or contrastive learning. It is also possible to make the model resistant to overfitting by improving its robustness. In addition, adversarial training is a potential direction.
  \item \textbf{Low-quality data.} Complementing attribute techniques can be used to address the challenge of missing information in graphs. Corrupted graph topology can be repaired by denoising, thus performing well in downstream tasks.
  \item \textbf{Over-smoothing.} For the over-smoothing problem, it is suggested to add a regularization term to the model, introduce subgraph information, or use decoupled representation learning.
  \item \textbf{1-WL.} It is recommended that considering the introduction of identity information or using higher-order structure feature around nodes, which can go beyond the 1-WL representation limitation.
\end{itemize}

\section{Open Issues}
\label{sec:open_issue}
Despite existing studies that have made progress in GAL, there still remain many open issues in terms of heterogeneity, spatio-temporal network, scalability, generalization.

\subsection{Heterogeneity}
Most graph augmentation learning methods focus on homogeneous graphs including augmenting nodes, structural attributes, or models. However, the continuous development of graph learning methods is still difficult to handle the problem of heterogeneity in graph data. Heterogeneous graphs contain a lot more information than homogeneous graphs. Most graphs are heterogeneous such as knowledge graphs and citation networks, so there is of great significance to handle heterogeneity. At present, there is a lack of studies about GAL on heterogeneous graphs. Due to the diversity of types of nodes and edges in heterogeneous graphs, it will be more difficult to study GAL than that on homogeneous graphs. The complexity problems brought by various attributes and the design of augmenting model should be paid more attention to.

\subsection{Spatio-temporal Dynamics}
Spatio-temporal network is a common graph data type in daily life, such as traffic networks. Data loss or data inaccuracy frequently occur in the process of data collection. The data loss and inaccuracy at one certain single time point will have a negative impact on the downstream task results. Spatio-temporal networks vary over time, the loss and inaccuracy happen every time. As a consequence, the increasing complexity of the data has a significant influence on GAL techniques. Moreover, spatio-temporal networks contain two aspects of time and space. How to balance the augmentation strategies of the two aspects and meanwhile design an effective model remain an open issue.

\subsection{Scalability}

Large-scale networks are ubiquitous in the real world, consisting of billions of nodes. As a result, graph learning models are generally time-consuming and have unaffordable space complexity. The corresponding problem definitely also exists in GAL, or even worse. In large-scale graphs, as the number of nodes or edges increases, the scale of augmentation data will also increase. But till now, there is no effective parallel solution for GAL. The problem of high cost, the selection of augmentation strategy, and the optimization of augmentation model are the main problems when facing scalability in GAL.

\subsection{Generalization}
Although GAL approaches have been proved to be effective in many different tasks, most approaches or strategies are designed for specific tasks or datasets. As it is known, graphs are very different in many domains, no matter in scale or structure. The consequence of this is that the generalization of the model is reduced. In GAL, the type and method of augmenting data will affect the effect of model generalization. If datasets and tasks are analyzed ahead of time, more overhead can be incurred. The generalization of the model still lacks theoretical exploration, and the stability of the algorithm also plays an important role in generalization.


\section{Conclusion}
\label{sec:conclusion}


Graph Augmentation Learning (GAL) techniques have received a lot of attention for their outperformance in different tasks. This survey provides a comprehensive overview of GAL techniques. Augmentation strategies are divided into macro, meso, and micro levels, respectively. The survey also summarizes GAL's ability in enhancing the robustness of graph learning models. Specifically, the outperformance of GAL is discussed. Three different situations namely low-quality data, model limitation, and hybrid are analyzed. In addition, this survey demonstrates the positive effects of GAL technologies through comparative experiments. Moreover, guidelines of suitable and feasible GAL strategies for different tasks are also provided. To the best of our knowledge, this survey paper is the first to discuss GAL techniques from such a systematical, comprehensive, and verifiable view.

GAL is a flourishing research field, which still remains many open issues as we have discussed in this survey paper, such as heterogeneity, dynamics, scalability, generalization, etc. Compared with other Euclidean data, the complexity of graph data grows exponentially faster. Low-quality graph data, such as the missing of relations or nodes within graph data, will have a much worse impact on downstream task accuracy than other Euclidean data. Therefore, the significance of GAL should be highly focused, and more efforts should be contributed to GAL as well.

\section{Acknowledgments}
This work is partially supported by the National Key Research and Development Program of China under Grant No. 2021ZD0112400, and National Natural Science Foundation of China under Grant No. 62102060. The authors would like to thank Yin Peng for his help with the experiments.

\bibliographystyle{unsrtnat}
\bibliography{main}  






\end{document}